\documentclass{article}

\usepackage[final]{nips_2016}

\usepackage[utf8]{inputenc} 
\usepackage[T1]{fontenc}    
\usepackage{hyperref}       
\usepackage{url}            
\usepackage{booktabs}       
\usepackage{amsfonts}       
\usepackage{nicefrac}       
\usepackage{microtype}      
\usepackage{amsmath}
\usepackage{epsfig}
\usepackage{epstopdf}
\usepackage{subfig,caption}
\usepackage{amssymb}

\title{Learning to Poke by Poking: Experiential Learning of Intuitive Physics}

\author{Pulkit Agrawal$^{*}$ \hspace{5mm} Ashvin Nair\thanks{equal contribution, authors are listed in alphabetical order.} \hspace{0.3mm} \hspace{5mm}  Pieter Abbeel \hspace{5mm} Jitendra Malik \hspace{5mm} Sergey Levine \\
Berkeley Artificial Intelligence Research Laboratory (BAIR) \\
University of California Berkeley \\
\texttt{\{pulkitag,anair17,pabbeel,malik,svlevine\}@berkeley.edu}
} 

\begin{document}

\maketitle

\begin{abstract}
We investigate an experiential learning paradigm for acquiring an internal model of intuitive physics. Our model is evaluated on a real-world robotic manipulation task that requires displacing objects to target locations by poking. The robot gathered over 400 hours of experience by executing more than 100K pokes on different objects. We propose a novel approach based on deep neural networks for modeling the dynamics of robot's interactions directly from images, by jointly estimating forward and inverse models of dynamics. The inverse model objective provides supervision to construct informative visual features, which the forward model can then predict and in turn regularize the feature space for the inverse model. The interplay between these two objectives creates useful, accurate models that can then be used for multi-step decision making. This formulation has the additional benefit that it is possible to learn forward models in an abstract feature space and thus alleviate the need of predicting pixels. Our experiments show that this joint modeling approach outperforms alternative methods. 
\end{abstract}

\section{Introduction}
\label{sec:intro}
Humans can effortlessly manipulate previously unseen objects in novel ways. 
For example, if a hammer is not available, a human might use a piece of rock or back of a screwdriver to hit a nail. What enables humans to easily perform such tasks that machines struggle with? One possibility is that humans possess an internal model of physics (i.e. ``intuitive physics'' \citep{michotte1963perception, mccloskey1983intuitive}) that allows them to reason about physical properties of objects and forecast their dynamics under the effect of applied forces. Such models can be used to transform a given task into a search problem in a manner similar to how moves can be planned in a game of chess or tic-tac-toe by searching through the game tree. Because the search algorithm is independent of task semantics, solutions to different and possibly new tasks can be determined using the same mechanism.

\begin{figure}
\begin{center}
\includegraphics[width=1.0\linewidth]{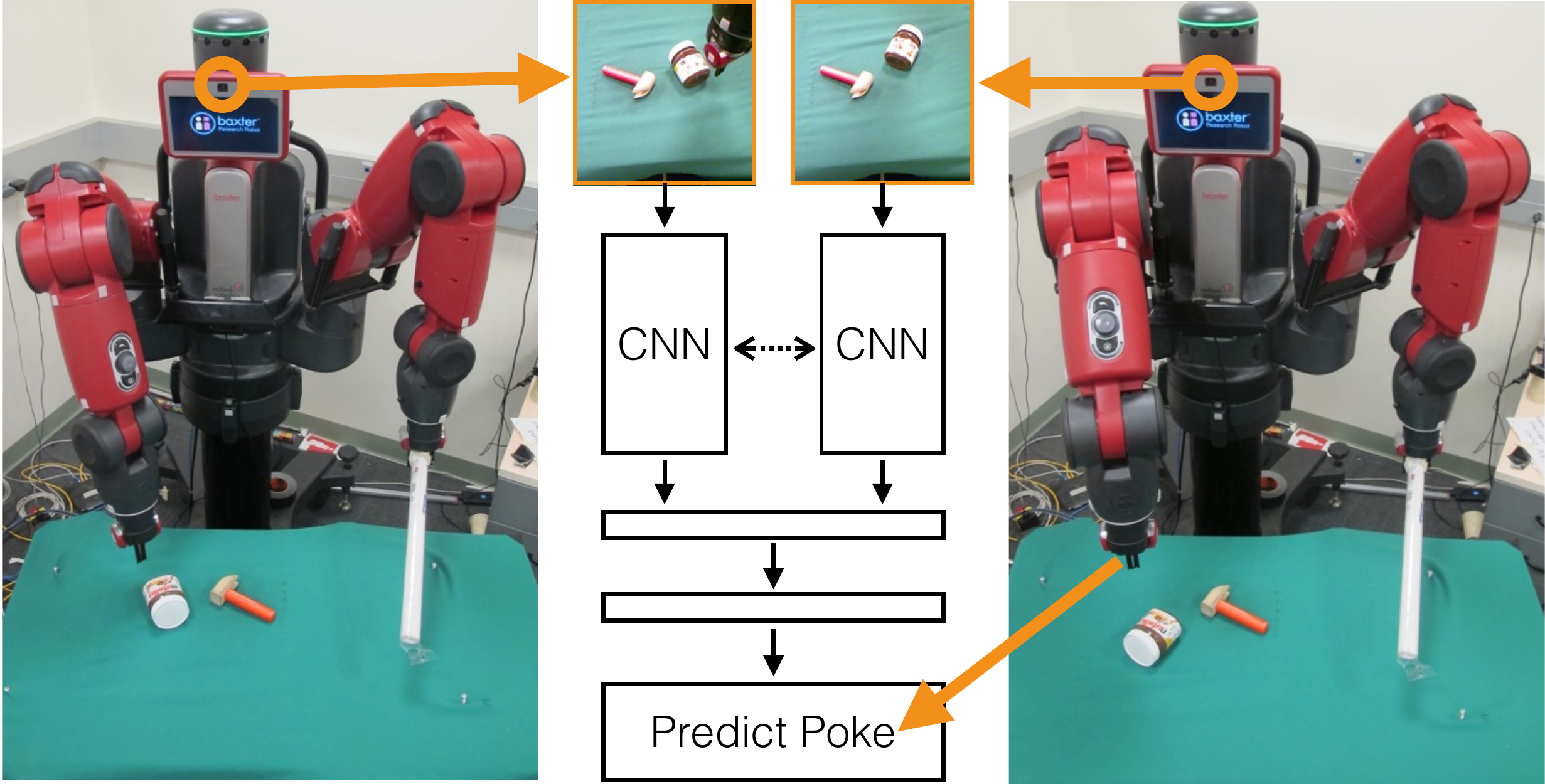}
\end{center}
\caption{Infants spend years worth of time playing with objects in a seemingly random manner. They might use this experience to learn a model of physics relating their actions with the resulting motion of objects. Inspired by this hypothesis, we let a robot interact with objects by randomly poking them. The robot pokes objects and records the visual state before (left) and after (right) the poke. The triplet of before image, after image and the applied poke is used to train a neural network (center) for learning the mapping between actions and the accompanying change in visual state. We show that this learn model can be used to push objects into a desired configuration.}
\label{fig:fig1}
\end{figure}

In human development, it is well known that infants spend years worth of time playing with objects in a seemingly random manner with no specific end goal 
\citep{smith2005development, gopnik1999scientist}. One hypothesis is that infants distill this experience into intuitive physics models that predict how their actions effect the motion of objects. Once learnt, these models could be used for planning actions for achieving novel goals later in life. Inspired by this hypothesis, in this work we investigate whether a robot can use it's own experience to learn an intuitive model of physics that is also effective for planning actions. In our setup (see Figure \ref{fig:fig1}), a Baxter robot interacts with objects kept on a table in front of it by randomly poking them. The robot records the visual state of the world before and after it executes a poke in order to learn a mapping between its actions and the accompanying change in visual state caused by object motion. To date our robot has interacted with objects for more than 400 hours and in process collected more than 100K pokes on 16 distinct objects. 

What kind of a model should the robot learn from it's experience? One possibility is to build a model that predicts the next visual state from the current visual state and the applied force (i.e forward dynamics model). This is challenging because predicting the value of every pixel in the next image is non-trivial in real world scenarios. Moreover, in most cases it is not the precise pixel values that are of interest, but the occurrence of a more abstract event. For example, predicting that a glass jar will break when pushed from the table onto the ground is of greater interest (and easier) than predicting exactly how every piece of shattered glass will look. The difficulty, however, is that supervision for such abstract concepts or events is not readily available in unsupervised settings such as ours. In this work, we propose one solution to this problem by jointly training forward and inverse dynamics models. A forward model predicts the next state from the current state and action, and an inverse model predicts the action given the initial and target state. In joint training, the inverse model objective provides supervision for transforming image pixels into an abstract feature space, which the forward model can then predict. The inverse model alleviates the need for the forward model to make predictions in the pixel space and the forward model in turn regularizes the feature space for the inverse model.  

We empirically show that the joint model allows the robot to generalize and plan actions for achieving tasks with significantly different visual statistics as compared to the data used in the learning phase. Our model can be used for multi step decision making and displace objects with novel geometry and texture into desired goal locations that are much farther apart as compared to position of objects before and after a single poke. We probe the joint modeling approach further using simulation studies and show that the forward model regularizes the inverse model. 

\section{Data}
Figure 1 shows our experimental setup. The robot is equipped with a Kinect camera and a gripper for poking objects kept on a table in front of it. At any given time there were 1-3 objects chosen from a set of 16 distinct objects present on the table. The robot's coordinate system was as following: X and Y axis represented the horizontal and vertical axes, while the Z axis pointed away from the robot. The robot poked objects by moving its finger along the XZ plane at a fixed height from the table. 

\textbf{Poke Representation:} For collecting a sample of interaction data, the robot first selects a random target point in its field of view to poke. One issue with random poking is that most pokes are executed in free space which severely slows down collection of interesting interaction data. For speedy data collection, a point cloud from the Kinect depth camera was used to only chose points that lie on any object except the table. Point cloud information was only used during data collection and at test time our system only requires RGB image data. After selecting a random point to poke ($p$) on the object, the robot randomly samples a poke direction ($\theta$) and length ($l$). Kinematically, the poke is defined by points $p_1, p_2$ that are $\frac{l}{2}$ distance from $p$ in the directions $\theta^{o}, (180 + \theta)^{o}$ respectively. The robot executes the poke by moving its finger from $p_1$ to $p_2$. 

\begin{figure}
\begin{center}
\includegraphics[width=0.75\linewidth]{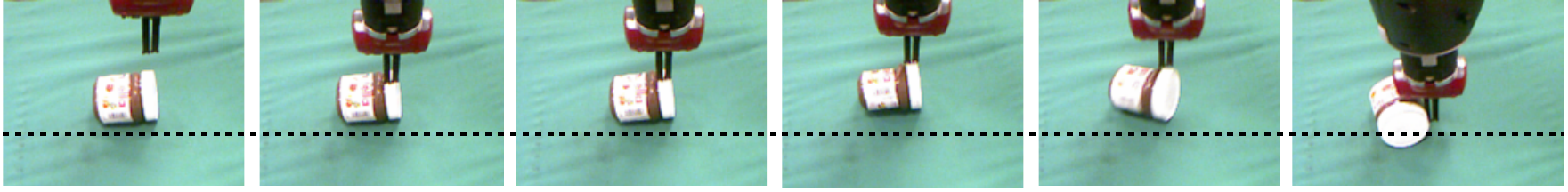}
\end{center}
\caption{These images depict the robot in the process of displacing the bottle away from the indicated dotted line. In the middle of the poke, the object flips and ends up moving in the wrong direction. Such occurrences are common because the real world objects have complex geometric and material properties. This makes learning manipulation strategies without prior knowledge very challenging.}
\label{fig:weird-seq}
\vspace{-0.1in}
\end{figure}

Our robot can run autonomously 24x7 without any human intervention. Sometimes when objects are poked they move as expected, but other times due to non-linear interaction between the robot's finger and the object they move in unexpected ways as shown in Figure \ref{fig:weird-seq}. Any model of the poking data must deal with such non-linear interactions (see \href{http://ashvin.me/pokebot-website/}{project website} for more examples).  A small amount of data in the early stages of the project was collected on a table with a green background, but most of our data was collected in a wooden arena with walls for preventing objects from falling down. All results in this paper are from data collected only from the wooden arena. 

\section{Method}
\label{sec:method}

The forward and inverse models can be formally described by equations \ref{eq:fwd} and \ref{eq:inv}, respectively. The notation is as following: $x_t, u_t$ are the world state and action applied time step $t$,  $\hat{x}_{t+1}, \hat{u}_{t+1}$ are the predicted state and actions, and $W_{fwd}$ and $W_{inv}$ are parameters of the functions $F$ and $G$ that are used to construct the forward and inverse models. 

\noindent\begin{minipage}{0.5\linewidth}
\begin{equation}
 \hat{x}_{t+1} = F(x_t, u_t; W_{fwd}) \label{eq:fwd}
\end{equation}
\end{minipage}%
\begin{minipage}{0.5\linewidth}
\begin{equation}
 \hat{u_t} = G(x_t, x_{t+1}; W_{inv}) \label{eq:inv}
\end{equation}
\end{minipage}\par\vspace{\belowdisplayskip}

Given an initial and goal state, inverse models provide a direct mapping to actions required for achieving the goal state in one step (if feasible). However, multiple possible actions can transform the world from one visual state to another. For example, an object can appear in a certain part of the visual field if the agent moves or if the agent uses its arms to move the object. 
This multi-modality in the action space makes the learning hard. On the other hand, given $x_t$ and $u_t$, there exists a next state $x_{t+1}$ that is unique up to dynamics noise. This suggests that forward models might be easier to learn. However, learning forward models in image space is hard because predicting the value of each pixel in the future frames is a non-trivial problem with no known good solution. However, in most scenarios we are not interested in predicting every pixel, but predicting the occurrence of a more abstract event such as object motion, change in object pose etc. 


The ability to learn an abstract task relevant feature space should make it easier to learn a forward dynamics model. One possible approach is to learn a dynamics model in the feature representation of a higher layer of a deep neural network trained to perform image classification (say on ImageNet)
~\citep{vondrickanticipating}. However, this is not a general way of learning task relevant features and it is unclear whether features adept at object recognition are also optimal for object manipulation. The alternative of adapting higher layer features of a neural network while simultaneously optimizing for the prediction loss leads to a degenerate solution of all the features reducing to zero, since the prediction loss in this case is also zero. Our key observation is that this degenerate solution can be avoided by imposing the constraint that it should be possible to infer the the executed action ($u_t$) from the feature representation of two images obtained before ($x_t$) and after ($x_{t+1}$) the action ($u_t$) is applied (i.e. optimizing the inverse model).  This formulation provides a general mechanism for using general purpose function approximators such as deep neural networks for simultaneously learning a task relevant feature space and forecasting the future outcome of actions in this learned space. 

A second challenge in using forward models is that inferring the optimal action inevitably leads to finding a solution to non-convex problems that are subject to local optima. The inverse model does not suffers from this drawback as it directly outputs the required action. These considerations suggest that inverse and forward models have complementary strengths and therefore it is worthwhile to investigate training a joint model of inverse and forward dynamics. 

\subsection{Model}
\label{sec:model}

\begin{figure}[t!]
\begin{center}
\includegraphics[width=1.0\linewidth]{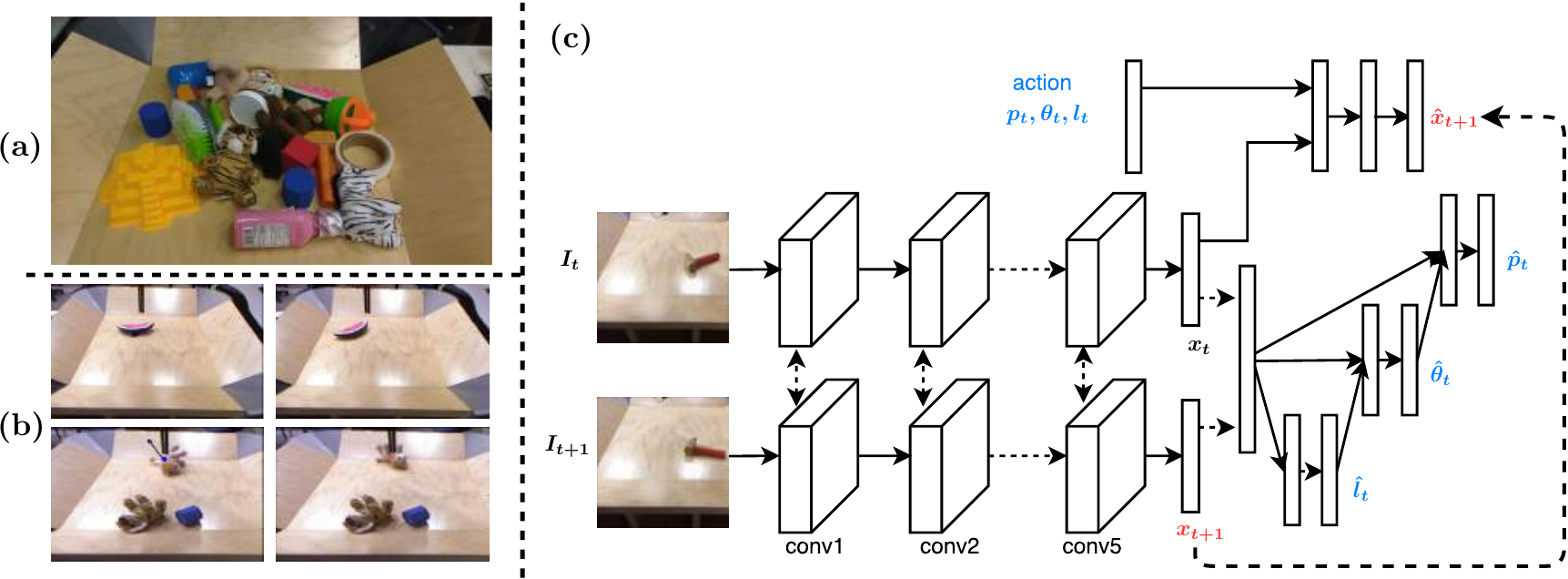}
\end{center}
\vspace{-0.1in}
\caption{(a) The collection of objects in the training set poked by the robot. (b) Example pairs of before ($I_{t}$) and after images ($I_{t+1}$) after a single poke was made by the robot. (c) A Siamese convolutional neural network was trained to predict the poke location ($p_t$), angle ($\theta_t$) and length ($l_t$) required to transform objects in the image at the $t^{th}$ time step ($I_{t}$) into their state in $I_{t+1}$. Images $I_{t}$ and $I_{t+1}$ are transformed into their latent feature representations ($x_{t}, x_{t+1}$) by passing them through a series of convolutional layers. For building the inverse model, $x_{t}, x_{t+1}$ are concatenated and passed through fully connected layers to predict the discretized poke. For building the forward model, the action $u_t$ = $\{p_t, \theta_t, l_t\}$ and $x_t$ are passed through a series of fully connected layers to predict $x_{t+1}$.}
\label{fig:method}
\vspace{-0.2in}
\end{figure}

A deep neural network is used to simultaneously learn a model of forward and inverse dynamics (see Figure \ref{fig:method}). A tuple of before image ($I_{t}$), after image ($I_{t+1}$) and the robot's action ($u_t$) constitute one training sample. Input images at consequent time steps ($I_t, I_{t+1}$) are transformed into their latent feature representations ($x_t, x_{t+1}$) by passing them through a series of five convolutional layers with the same architecture as the first five layers of AlexNet \citep{krizhevsky2012imagenet}. For building the inverse model, $x_{t}, x_{t+1}$ are concatenated and passed through fully connected layers to conditionally predict the poke location ($p_t$), angle ($\theta_t$) and length ($l_t$) separately. For modeling multimodal poke distributions, poke location, angle and length of poke are discretized into a $20\times 20$ grid, 36 bins and 11 bins respectively. The $11^{th}$ bin of the poke length is used to denote no poke. For building the forward model, the feature representation of the before image ($x_t$) and the action ($u_t$; real-valued vector without discretization) are passed into a sequence of fully connected layer that predicts the feature representation of the next image ($x_{t+1}$). Training is performed to optimize the loss defined in equation \ref{eq:joint} below. 
\begin{align}
\label{eq:joint}
     L_{joint} = L_{inv}(u_t, \hat{u_t}, W) \hspace{1mm} + & \hspace{1mm}  \lambda L_{fwd}(x_{t+1}, \hat{x}_{t+1}, W)
\end{align}
$L_{inv}$ is a sum of three cross entropy losses between the actual and predicted poke location, angle and length. $L_{fwd}$ is a L1 loss between the predicted ($\hat{x}_{t+1}$) and the ground truth ($x_{t+1}$) feature representation of the after image ($I_{t+1}$). $W$ are the parameters of the neural network. We used $\lambda = 0.1$ in all our experiments. We call this the joint model and we compare its performance against the inverse only model that was trained by setting $\lambda$ = 0 in equation \ref{eq:joint}. More details about model training are provided in the \href{http://ashvin.me/pokebot-website/}{supplementary materials}. 

\subsection{Evaluation Procedure}
\label{sec:evalproc}

\begin{figure}[t!]
\begin{center}
\includegraphics[width=1.0\linewidth]{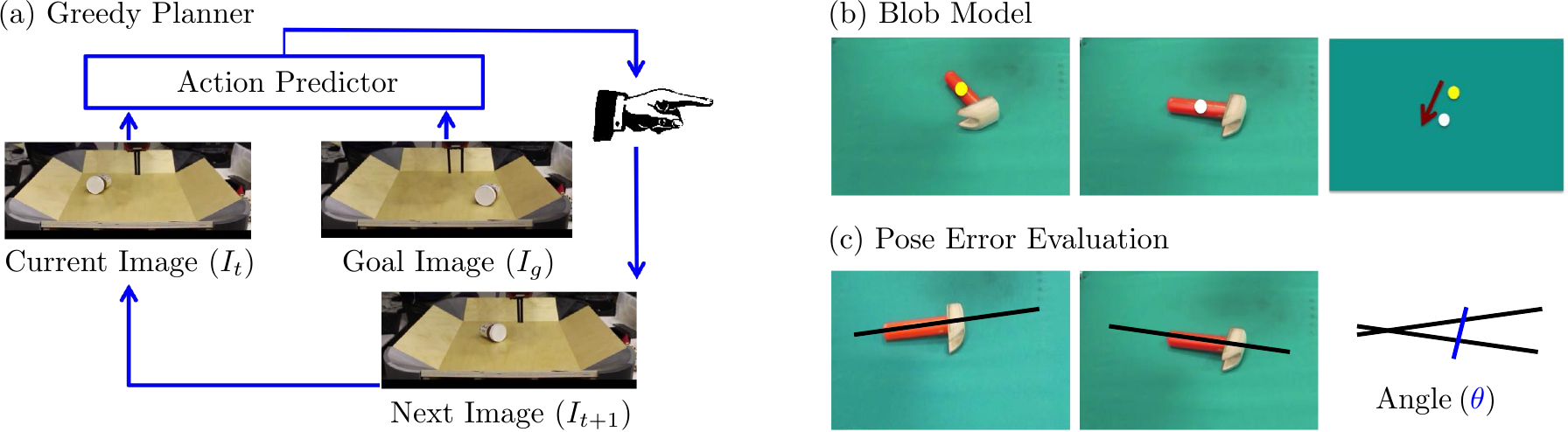}
\end{center}
\vspace{-0.1in}
\caption{(a) Greedy planner is used to output a sequence of pokes to displace the objects from their configuration in initial to the goal image. (b) The blob model first detects the location of objects in the current and goal image. Based on object positions, location and angle of the poke is computed and then executed by the robot. The obtained next and goal image are used to compute the next poke and this process is repeated iteratively. (c) The error of the models in poking objects to their correct pose is measured as the angle between the major axis of the objects in the final and goal images. }
\label{fig:planner}
\end{figure}

One way to test the learnt model is to provide the robot with an initial and goal image and task it to apply pokes that would displace objects into the configuration shown in the goal image. If the robot succeeds at achieving the goal configuration when the visual statistics of the pair of initial and goal image is similar to before and after image in the training set, then this would not be a convincing demonstration of generalization. However, if the robot is able to displace objects into goal positions that are much farther apart as compared to position of objects before and after a single poke then it might suggest that our model has not simply overfit but has learnt something about the underlying physics of how objects move when poked. This suggestion would be further strengthened if the robot is also able to push objects with novel geometry and texture in presence of multiple distractor objects. 

If the objects in the initial and goal image are farther apart than the maximum distance that can be pushed by a single poke, then the model would be required to output a sequence of pokes. We use a greedy planning method (see Figure \ref{fig:planner}(a)) to output a sequence of pokes. First, images depicting the initial and goal state are passed through the learnt model to predict the poke which is then executed by the robot. Then, the image depicting the current world state (i.e. the current image) and the goal image are fed again into the model to output a poke. This process is repeated iteratively unless the robot predicts a no-poke (see section \ref{sec:model}) or a maximum number of 10 pokes is reached. 

\textbf{Error Metrics:} In all our experiments, the initial and goal images differ in the position of only a single object. The location and pose of the object in the final image after the robot stops and the goal image are compared for quantitative evaluation. The location error is the Euclidean distance between the object locations. In order to account for different object distances in the initial and goal state, we use relative instead of absolute location error. Pose error is defined as the angle (in degrees) between the major axis of the objects in the final and goal images (see Figure \ref{fig:planner}(c)). Please see \href{http://ashvin.me/pokebot-website/}{supplementary materials} for further details. 

\subsection{Blob Model}
\label{sec:blob}
We compared the performance of the learnt model against a baseline blob model. This model first estimates object locations in current and goal image using template based object detector. It then uses the vector difference between these to compute the location, angle and length of poke executed by the robot (see \href{http://ashvin.me/pokebot-website/}{supplementary materials} for details). In a manner similar to greedy planning with the learnt model, this process is repeated iteratively until the object gets closer to the desired location in the goal image by a pre-defined threshold or a maximum number of pokes is reached.

\section{Results}
\label{sec:results}

\begin{figure}[t!]
\begin{center}
\includegraphics[width=1.0\linewidth]{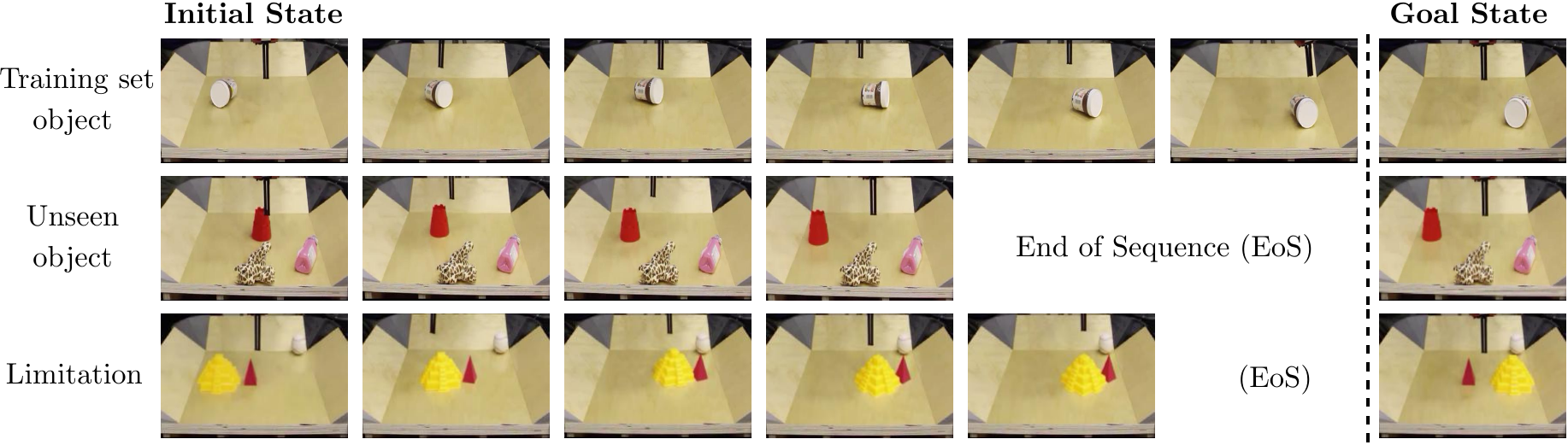}
\end{center}
\vspace{-0.1in}
\caption{The robot is able to successfully displace objects in the training set (row 1; Nutella bottle) and objects with previously unseen geometry (row 2; red cup) into goal locations that are significantly farther than pair of before and after images used in the training set. The robot is unable to push objects around obstacles (row 3;  limitation of greedy planning).}
\label{fig:poke_results}
\end{figure}

The robot was tasked to displace objects in an initial image into their configuration depicted in a goal image (see Figure \ref{fig:poke_results}). The three rows in the figure show the performance when the robot is asked to displace an object (Nutella bottle) present in the training set, an object (red cup) whose geometry is different from objects in the training set and when the task is to move an object around an obstacle. These examples are representative of the robot's performance and more examples can be found on the \href{http://ashvin.me/pokebot-website/}{project website}. It can be seen that the robot is able to successfully poke objects present in the training set and objects with novel geometry and texture into desired goal locations that are significantly farther than pair of before and after images used in the training set. 

Row 2 in Figure \ref{fig:poke_results} also shows that the robot's performance in unaffected by the presence of distractor objects that occupy the same location in the current and goal images. These results indicate that the learnt model allows the robot to perform tasks that show generalization beyond the training set (i.e. poking object by small distances). Row 3 in Figure \ref{fig:poke_results} depicts an example where the robots fails to push the object around an obstacle (yellow object). The robot acts greedily and ends up pushing the obstacle along with the object. One more side-effect of greedy planning is zig-zag instead of straight trajectories taken by the object between its initial and goal locations. Investigating alternatives to greedy planning, such as using the learnt forward model for planning pokes is a very interesting direction for future research. 

What representation could the robot have learnt that allows it to generalize? One possibility is that the robot ignores the geometry of the object and only infers the location of the object in the initial and goal image and uses the difference vector between object locations to deduce what poke to execute. This strategy is invariant to absolute distance between the object locations and is therefore capable of explaining the observed generalization to large distances. While we cannot prove that the model has learnt to detect object location, nearest neighbor visualizations of the learnt feature space clearly suggest sensitivity to object location (see  \href{http://ashvin.me/pokebot-website/}{supplementary materials}). This is interesting because the robot received no direct supervision to locate objects.  

Because different objects have different geometries, they need to be poked at different places to move them in the same manner. For example, a Nutella bottle can be reliably moved forward without rotating the bottle by poking it on the side along the direction toward its center of mass, whereas a hammer is reliably moved by poking it where the hammer head meets the handle. Pushing an object to a desired pose is harder and requires a more detailed understanding of object geometry in comparison to pushing the object to a desired location. In order to test whether the learnt model represents any information about object geometry, we compared its performance against the baseline blob model (see section \ref{sec:blob} and figure \ref{fig:planner}(b)) that ignores object geometry. For this comparison, the robot was tasked to push objects to a nearby goal by making only a single poke (see  \href{http://ashvin.me/pokebot-website/}{supplementary materials} for more details). Results in Figure \ref{fig:quantitative}(a) show that both the inverse and joint model outperform the blob model. This indicates that in addition to representing information about object location, the learn models also represent some information about object geometry.

\subsection{Forward model regularizes the inverse model}
\begin{figure}[t!]
\begin{center}
\includegraphics[width=1.0\linewidth]{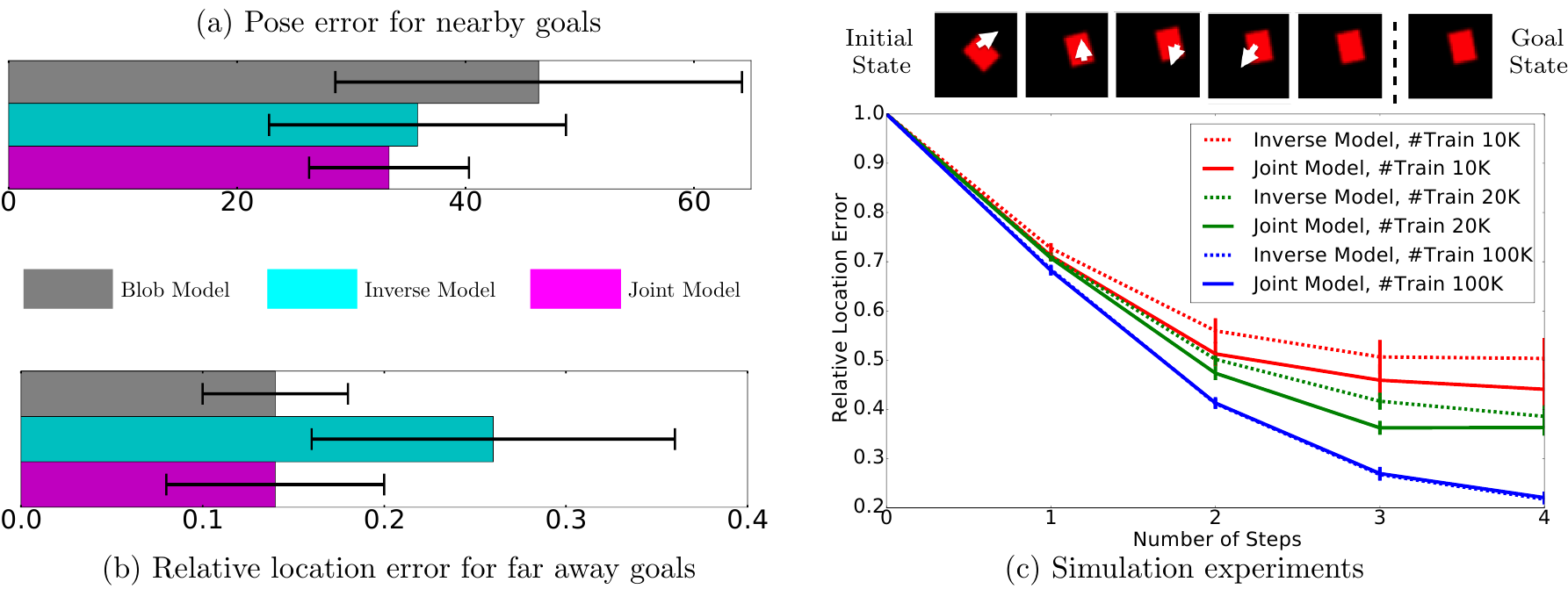}
\end{center}
\vspace{-0.1in}
\caption{(a) Inverse and Joint model are more accurate than the blob model at pushing objects towards the desired pose. (b) The joint model outperforms the inverse-only model when the robot is tasked to push objects by distances that are significantly larger than object distance in before and after images used in the training set (i.e. a test of generalization). (c) Simulation studies reveal that when less number of training examples (10K, 20K) are available the joint model outperforms the inverse model and the performance is comparable with larger amount of data (100K). This result indicates that the forward model regularizes the inverse model. } 
\label{fig:quantitative}
\end{figure}

We tested the hypothesis whether the forward model regularizes the feature space learnt by the inverse model in a 2-D simulation environment where the agent interacted with a red rectangular object by poking it by small forces. The rectangle was allowed to freely translate and rotate (Figure \ref{fig:quantitative}(c)). Model training was performed using an architecture similar to the one described in section \ref{sec:model}.  Additional details about the experimental setup, network architecture and training procedure for the simulation experiments are provided in the \href{http://ashvin.me/pokebot-website/}{supplementary materials}. Figure \ref{fig:quantitative}(c) shows that when less training data (10K, 20K examples) is available the joint model outperforms the inverse model and reaches closer to the goal state in fewer steps (i.e. fewer actions). This shows that indeed the forward model regularizes the inverse model and helps generalize better. However, when the number of training examples is increased to 100K both models are at par. This is not surprising because training with more data often results in better generalization and thus the inverse model is no longer reliant on the forward model for the regularization. 

Evaluation on the real robot supports the findings from the simulation experiments. Figure \ref{fig:quantitative}(b) shows that in a test of generalization, when an object is required to be displaced by a long distance, the joint model outperforms the inverse model. Similar performance of joint and blob model at this task is not surprising because even if the pokes are somewhat inaccurate but generally in the direction from object's current to goal location, the object might traverse a zig-zag path but it would eventually reach the goal. The joint model is however more accurate at displacing objects into their correct pose as compared to the blob model (Figure \ref{fig:quantitative}(a)).  




\section{Related Work}


Learning visual control policies using reinforcement learning for tasks such as playing Atari games~\citep{mnih2015human}, controlling robots in simulation~\citep{lillicrap2015continuous} and in the real world~\citep{LevineFDA15} is of growing interest.
However, these methods are model free and learn goal specific policies, which makes it difficult to repurpose the learned policies for new tasks. In contrast, the aim of this work is to learn intuitive physical models of object interaction which we show allow the agent to generalize. 
Other works in visual control  have relied on model free methods that operate on a a low-dimensional state representation of images obtained using autoencoders \citep{lange2012autonomous, finn2015deep,kietzmann2009neuro}. It is unclear that features obtained by optimizing pixelwise reconstruction are necessarily well suited for model based control.

Learning to grasp objects by trial and error from large amounts of interaction data has recently been explored \citep{pinto2015supersizing,levine2016learning}. These methods aim to acquire a policy for solving a single concrete task, while our work is concerned with learning a general predictive model that could be used to achieve a variety of goals at test time. When an object is grasped, it is possible to fully control the state of the grasped object. However, in non-prehensile manipulation (i.e. manipulation without grasping \citep{lavalle2006planning}) such as poking, the object state is not directly controllable which makes manipulation by poking harder than grasping \citep{dogar2012planning}. Learning a model of poking was considered by \citep{pinto2016curious}, but their goal was to learn visual representations and they did not consider using the learnt models to displace objects to goal locations. 


A good review of model based control can be found in \citep{mayne2014model} and \citep{jordan1992forward,wolpert1995internal} provide interesting perspectives. A model based deep learning method for cutting vegetables was considered by \citep{lenz2015deepMPC}. However, as their system operated on the robotic state space instead of vision and is thus limited in its generality.  Model based control from visual inputs was considered by \citep{billiards,wahlstrom2015from,watter2015embed,actionconditioned}  in synthetic domains of manipulating two degree of freedom robotic arm, inverted pendulum, billiards and Atari games. In contrast, we tackle manipulation of complex, compressible real world objects. Instead of learning a model of physics, some recents works ~\citep{wu2015galileo, mottaghi2015newtonian, lerer2016learning} have proposed to use Newtonian physics in combination with neural networks to forecast object dynamics. 

In robotic manipulation, a number of prior methods have been proposed that use hand-designed visual features and known object poses or key locations to plan and execute pushes and other non-prehensile manipulations \citep{kopicki2011learning,lau2011automatic,mericcli2015push}. Unlike these methods, the goal in our work is to learn an intuitive physics model for pushing only from raw images, thus allowing the robot to learn by exploring the environment on its own without human intervention.

\section{Discussion and Future Work}



In this work we propose to learn ``intuitive" model of physics using interaction data. An alternative is to represent the world in terms of a fixed set of physical parameters such as mass, friction coefficient, normal forces etc and use a physics simulator for computing object dynamics from this representation \citep{kolev2015physically, mottaghi2015newtonian, wu2015galileo,  hamrick2011internal}. This approach is general because physics simulators inevitably use Newton's laws that apply to a wide range of physical phenomenon ranging from orbital motion of planets to a swinging pendulum. Estimating parameters such as as mass, friction coefficient etc. from sensory data is subject to errors, and it is possible that one parameterization is easier to estimate or more robust to sensory noise than another. For example, the conclusion that objects with feather like appearance fall slower than objects with stone like appearance can be reached by either correlating visual texture to the speed of falling objects, or by computing the drag force after estimating the cross section area of the object. Depending on whether estimation of visual texture or cross section area is more robust, one parameterization will result in more accurate predictions than the other. Pre-defining a set of parameters for predicting object dynamics, which is required by ``simulator-based" approach might therefore lead to suboptimal solutions that are less robust. 

For many practical object manipulation tasks of interest, such as re-arranging objects, cutting vegetables, folding clothes, and so forth, small errors in execution are acceptable. The key challenge is robust performance in the face of varying environmental conditions. This suggests that a more robust but a somewhat imprecise model may in fact be desirable over a less robust and a more precise model. While the arguments presented above suggest that intuitive physics models are likely to be more robust than simulator based models, quantifying the robustness of these models is an interesting direction for future work. Furthermore, it is non-trivial to use simulator based models for manipulating deformable objects such as clothes and ropes because simulation of deformable objects is hard and also also requires representing objects by heavily handcrafted features that are unlikely to generalize across objects. The intuitive physics approach does not make any object specific assumptions and can be easily extended to work with deformable objects. This approach is in the spirit of recent successful deep learning techniques in computer vision and speech processing that learn features directly from data, whereas the simulator based physics approach is more similar to using hand-designed features. Current methods for learning intuitive physics models, such as ours are data inefficient and it is possible that combining intuitive and simulator based approaches leads to better models than either approach by itself.

In poking based interaction, the robot does not have full control of the object state which makes it harder to predict and plan for the outcome of an action. The models proposed in this work generalize and are able to push objects into their desired location. However, performance on setting objects in the desired pose is not satisfactory, possibly because of the robot only executing pokes in large, discrete time steps. An interesting area of future investigation is to use continuous time control with smaller pokes that are likely to be more predictable than the large pokes used in this work. Further, although our approach is evaluated on a specific robotic manipulation task, there are no task specific assumptions, and the techniques are applicable to other tasks. In future, it would be interesting to see how the proposed approach scales with more complex environments, diverse object collections, different manipulation skills and to other non-manipulation based tasks, such as navigation. Other directions for future investigation include the use of forward model for planning and developing better strategies for data collection than random interaction.

\textbf{Supplementary Materials:} and videos can be found at \href{http://ashvin.me/pokebot-website/}{http:\//\//ashvin.me\//pokebot-website\//}.

\textbf{Acknowledgement:} We thank Alyosha Efros for inspiration and fruitful discussions throughout this work. The title of this paper is partly influenced by the term ``pokebot" that Alyosha has been using for several years. We thank Ruzena Bajcsy for access to Baxter robot and Shubham Tulsiani for helpful comments. This work was supported in part by ONR MURI N00014-14-1-0671, ONR YIP and by ARL through the MAST program. We are grateful to NVIDIA corporation for donating K40 GPUs and providing access to the NVIDIA PSG cluster.

\let\OLDthebibliography\thebibliography
\renewcommand\thebibliography[1]{
	\OLDthebibliography{#1}
	\setlength{\itemsep}{5pt plus 0.3ex}
}

{\scriptsize
\bibliographystyle{icml}
\bibliography{main_arxiv_v2}
}
    
\end{document}